\documentclass[10pt,twocolumn,letterpaper]{article}

\usepackage{cvpr}
\usepackage{times}
\usepackage{epsfig}
\usepackage{graphicx}
\usepackage{amssymb}
\usepackage{helvet}  
\usepackage{courier}  
\usepackage{url}  
\usepackage{graphicx}  
\usepackage{multirow,floatrow, comment, enumerate}
\usepackage{amsmath,amssymb}
\usepackage{color}
\usepackage{algorithmic}
\usepackage[linesnumbered,ruled,vlined]{algorithm2e}
\usepackage{subcaption}
\usepackage[pagebackref=true,breaklinks=true,letterpaper=true,colorlinks,bookmarks=false]{hyperref}

\cvprfinalcopy 

\def\cvprPaperID{} 

\begin{document}

\title{Adaptation and Re-Identification Network:\\
An Unsupervised Deep Transfer Learning Approach to Person Re-Identification}

\author{Yu-Jhe Li$^{1}$, Fu-En Yang$^{2}$, Yen-Cheng Liu$^{1}$, Yu-Ying Yeh$^{1}$, Xiaofei Du$^{3}$, Yu-Chiang Frank Wang$^{1,2}$\\
$^1$Graduate Institute of Communication Engineering, National Taiwan University\\
$^2$Department of Electrical Engineering, National Taiwan University\\
$^3$Umbo Computer Vision\\
{\tt\small \{r06942074, b03801039, r04921003, b99202023\}@ntu.edu.tw, }\\
{\tt\small xiaofei.du@umbocv.com, ycwang@ntu.edu.tw}
}

\maketitle

\begin{abstract}
Person re-identification (Re-ID) aims at recognizing the same person from images taken across different cameras. To address this task, one typically requires a large amount labeled data for training an effective Re-ID model, which might not be practical for real-world applications. To alleviate this limitation, we choose to exploit a sufficient amount of pre-existing labeled data from a different (auxiliary) dataset. By jointly considering such an auxiliary dataset and the dataset of interest (but without label information), our proposed adaptation and re-identification network (ARN) performs unsupervised domain adaptation, which leverages information across datasets and derives domain-invariant features for Re-ID purposes. In our experiments, we verify that our network performs favorably against state-of-the-art unsupervised Re-ID approaches, and even outperforms a number of baseline Re-ID methods which require fully supervised data for training.
\end{abstract}

\section{Introduction} \label{sec:intro}
Person re-identification (Re-ID)~\cite{zheng2016person}  has become popular research topic due to its application to smart city and large-scale surveillance system.
Given a person-of-interest (query) image, Re-ID aims at associating the same pedestrian from multiple cameras, matching people across non-overlapping camera views.
Yet, current Re-ID models are still struggling to handle the problems with intensive changes in appearance and environment.
With recent advances in deep neural networks, several works have been proposed to tackle the above challenges in supervised~\cite{sun2017svdnet, cheng2016person, lin2017improving, hermans2017defense,zhong2017camera, si2018dual} and unsupervised manners~\cite{wang2016towards,fan2017unsupervised, peng2016unsupervised, yu2017cross}.

\begin{figure}[t!]
	\centering
	\includegraphics[width=1.0\textwidth]{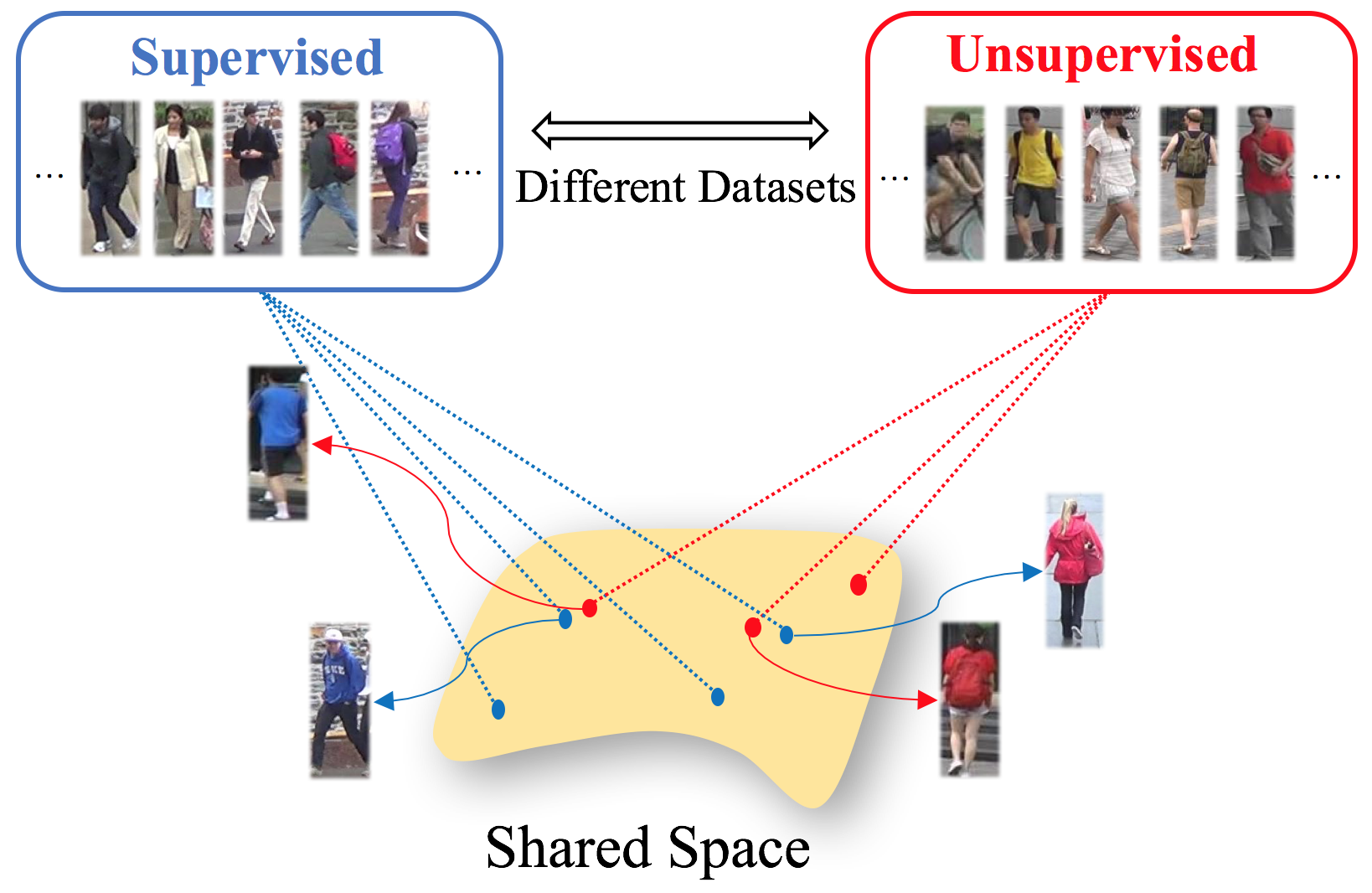}
     \caption{ 
Illustration of cross-dataset person re-identification (Re-ID). While Re-ID of images in the target-domain dataset is of interest, no labeled data is available for training. Our idea is to leverage information from auxiliary labeled images in a distinct and irrelevant source domain (i.e., dataset not of interest). With such an unsupervised domain adaptation setting for learning domain-invariant features, Re-ID in the target domain can be performed accordingly.
}
	\vspace{0mm}
	\label{fig:introduction}
\end{figure}

However, the aforementioned methods are not able to achieve satisfactory performances if the appearance or camera settings of query images are very different from the training ones. 
This is known as the problem of \textit{domain shift} (or \textit{domain/dataset bias}) and requires domain adaptation~\cite{patel2015visual} techniques to address this challenging yet practical problem. 
Thus, several works~\cite{zhong2017camera,image-image18} have been proposed to generalize the discriminative ability across different datasets by increasing the cross-domain training samples with style transfer methods.
Zhong \textit{et al.}~\cite{zhong2017camera} smooth style disparities across the cameras with style transfer model and label smooth regularization. Similarly, Deng \textit{et al.}~\cite{image-image18} further add similarity constraints to enhance the performance on cross-domain Re-ID task. 
However, the adaptation models based on style transfer are not necessary to preserve the identity during the image translation procedure, and this results in unsatisfactory performance when no corresponding identities appear in both domains/datasets.

To address the domain shifts between datasets, we propose a deep architecture to perform cross-domain Person Re-identification with the only supervision from a single dataset/domain as shown in Figure~\ref{fig:introduction}. Toward this end, with the labeled data, our model derives the discriminative property to distinguish the images between different classes.
To perform such property on alternative domain without annotation, our model learns to adapt the discriminative property from supervised (i.e., source) to unsupervised (i.e., target) domain. This is achieved by decomposing the cross-domain feature into domain-invariant and domain-specific one. Once the domain-invariant feature is learned, our model can perform cross-domain Re-ID by matching the query image and gallery images in the shared latent space.
To further enhance the discriminative property of our proposed model, we aim at increasing the margin between the classes with our proposed contrastive objective, which is later verified in the experiment.

The contributions of our paper can be summarized as follows:

 \begin{itemize}
 \item We address unsupervised person Re-ID by exploiting and adapting information learned from an auxiliary labeled dataset, which can be viewed as a unsupervised domain adaptation approach.
 \item Our proposed Adaptation and Re-ID Network (ARN) aims at learning domain-invariant features for matching images of the same person, while no label information is required for the data domain of interest.
  \item Our ARN not only performs favorably agianst state-of-the-art Re-ID approaches in the unsupervised setting, it also outperforms baseline supervised Re-ID methods.
\end{itemize}

\section{Related Works} \label{sec:related}

\noindent\textbf{2.1. Person Re-Identification (Re-ID) }

\textbf{Supervised Learning for Re-ID:}
Most existing Re-ID models are learned in a supervised setting. That is, given a sufficient number of labeled images across cameras, techniques based on metric learning~\cite{cheng2016person} or representation learning~\cite{lin2017improving} can be applied to train the associated models. {Cheng \textit{et al.}~\cite{cheng2016person}} propose a multi-channel part-based convolutional network for Re-ID, which is formulated via an improved triplet framework. {Lin \textit{et al.}}~\cite{lin2017improving} present an attribute-person
recognition network which performs discriminative embedding for Re-ID and is able to make a prediction for particular attributes. While promising performances have been reported on recent datasets (e.g., Market-1501~\cite{zheng2015scalable}, DukeMTMC-ReID~\cite{zheng2017unlabeled}), it might not be practical since collecting a large amount of annotated training data is typically computationally prohibitive.

\textbf{Unsupervised Learning for Re-ID:}
To alleviate the above limitation, researchers also focus on person Re-ID using unlabeled training data~\cite{fan2017unsupervised,wang2016towards}. For example, {Fan \textit{et al.}}~\cite{fan2017unsupervised} apply techniques of data clustering, instance selection, and fine-tuning methods to obtain pseudo labels for the unlabeled data; this allows the training of the associated feature extractor with discriminative ability. {Wang \textit{et al.}}~\cite{wang2016towards} propose a kernel-based model to learn cross-view identity discriminative information from unlabeled data. Nevertheless, due to the lack of label information for images across cameras, unsupervised learning based methods typically cannot achieve comparable results as the supervised approaches do.

\noindent\textbf{2.2. Cross-Domain Re-ID}

Recently, some transfer learning algorithms~\cite{geng2016deep,peng2016unsupervised} are proposed to leverage the Re-ID models pre-trained in source datasets to improve the performance on target dataset. {Geng \textit{et al.}}~\cite{geng2016deep} transfer representations learned from large image classification datasets to Re-ID datasets using a deep neural network which combines classification loss with verification loss. {Peng \textit{et al.}}~\cite{peng2016unsupervised} propose a multi-task dictionary learning model to transfer a view-invariant representation from a labeled source dataset to an unlabeled target dataset.

Besides, domain adaption and image--to--image translation approaches have been applied to Re-ID tasks increasingly, {Deng \textit{et al.}~\cite{image-image18}} combine CycleGAN~\cite{CycleGAN2017} with similarity constraint for domain adaptation which improve performance in cross-dataset setting. {Zhong \textit{et al.}}~\cite{zhong2017camera} introduce camera style transfer approach to address image style variation across multiple views and learn a camera-invariant descriptor subspace.

\noindent\textbf{2.3. Domain-Invariant Feature Learning}

We deal with the cross-domain Re-ID by learning domain-invariant feature. Here we review the recent works~\cite{tzeng2014deep,bousmalis2016domain,liu2016coupled,liu2017unsupervised} on learning domain-invariant feature.
In order to achieve cross-domain classification tasks, Tzeng \textit{et al.}~\cite{tzeng2014deep} present domain confusion loss to learn domain-invariant representation. Bousmalis \textit{et al.}~\cite{bousmalis2016domain} propose to extract the domain-invariant feature to improve the performance of cross-domain classification task. On the other hand, to tackle the problem of image style translation, Coupled GAN~\cite{liu2016coupled} also learn to synthesize cross-domain images from a domain-invariant feature. UNIT~\cite{liu2017unsupervised} further learn a domain-invariant feature to translate the image across domains.
\textcolor{black}{It is worth noting that, inspired by the above methods, we address the cross-domain Re-ID task by learning the domain-invariant feature for describing the human identity across distinct domains.}

\begin{figure*}[t!]
	\centering
	\includegraphics[width=0.99\textwidth]{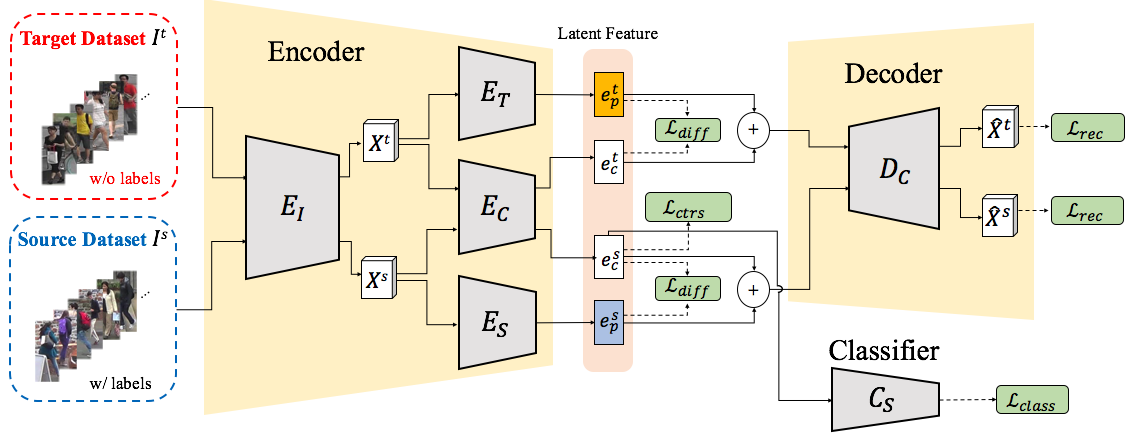}
    \caption{The architecture of our Adaptation and Re-Identification Network (ARN). Note that the Encoder contains the two shared modules ($E_I$, $E_C$), and two private modules ($E_T$, $E_S$). $E_I$ aims to retrieve visual feature maps ($X^t, X^s$), which are fed into $E_C$, $E_S$, and $E_T$ for learning domain-invariant (shared) and specific (private) features. With the private ($e^t_p$, $e^s_p$) and shared ($e^t_c$, $e^s_c$) latent features observed, the Decoder $D_C$ performs feature reconstruction for both target and source-domain images. Finally, the classifier $C_S$ is designed to perform supervised learning from source-domain data.} 
	\label{fig:archi}
\end{figure*}

\section{Proposed Method}\label{sec:method}
Given a set of image-label pairs $\{I_i^s, y_i^s\}_{i=1}^{N_s}$ and another set of images $\{I_i^t\}_{i=1}^{N_t}$, where $N_s$ and $N_t$ denote the total images of source and target dataset respectively, the goal of our model is to perform cross-dataset Re-ID by adapting the discriminative ability learned from source dataset to unlabeled target dataset. 



We present our Person Re-ID model trained in a supervised manner in section~\ref{ssec:sup_reid}. To address the cross-dataset Person Re-ID, our model leverages information from supervised data and adapts it to unsupervised dataset in section~\ref{ssec:uda_reid}. Later in section~\ref{ssec:learn_eva}, we demonstrate the learning and evaluation of our proposed method.

\subsection{Supervised Learning for Person Re-ID}\label{ssec:sup_reid}
To perform person re-identification, our model aims to learn the image feature with discriminative property to distinguish between classes. With labeled data, such feature property can be learned from image classification task. To achieve this, we introduce encoder $\lbrace E_I, E_C\rbrace$ and classifier $C_S$ to extract the image feature $e^s_c$ from source dataset image $I^s$ and obtain its category prediction $\hat{y}^s$ respectively. 
Specifically, to reduce the training burden, pre-trained model (e.g., ResNet) can be used for feature extractor module $E_I$. Thus, we define the classification loss $\mathcal{L}_{class}$ to minimize the negative log-likelihood of the ground truth label $y^s$ for source dataset image $I^s$:
\begin{equation}\label{eq:euclidean}
\begin{aligned}
\mathcal{L}_{class} = -\sum_{i=1}^{N_s}y_i^s \cdot \log \hat{y}_i^s
\end{aligned}
\end{equation}

To further enhance the discriminative property of our learned feature, we consider contrastive loss $\mathcal{L}_{ctrs}$~\cite{hadsell2006dimensionality} as an additional objective of our model:
\small
\begin{equation}\label{eq:contrastive}
\begin{aligned}
\mathcal{L}_{ctrs}= \sum_{i,j} \lambda(e^s_{c,i} - e^s_{c,j})^2
+( 1-\lambda)[max(0, m - (e^s_{c,i} - e^s_{c,j})]^2
\end{aligned}
\end{equation}
\normalsize
where $\lambda=1$ if $\lbrace e^s_{c,i}, e^s_{c,j} \rbrace$ belong to same category, and $\lambda=0$ if $\lbrace e^s_{c,i}, e^s_{c,j} \rbrace$ belong to different categories. Note that $m > 0$ is a margin, which is regarded a radius around $E_c(x_i)$. Dissimilar pairs contribute to the loss function only if their distance is within this radius.

However, the above supervised model cannot be directly applied to alternative dataset without label annotation. Thus, we further consider the adaption technique to generalize the discriminative ability to dataset without any label annotation.

\subsection{Unsupervised Domain Adaptation for Cross-Dataset Re-ID}\label{ssec:uda_reid}

Here we regard cross-dataset Re-ID as adaptation of discriminative ability from supervised source dataset to unsupervised target dataset. To this end, our model aims to eliminate dataset shift in the procedure of inferencing discriminative feature. 
Thus, as depicted in Figure~\ref{fig:archi}, our model first introduces $E_S$/$E_T$ to decompose visual feature maps $X^s$/$X^t$ into dataset-invariant feature $e^s_c$/$e^t_c$ and dataset-specific feature $e^s_p$/$e^t_p$. Our model acquires discriminative ability by applying the dataset-invariant feature $e^s_c$ to predict its corresponding category. Once such feature is learned, even without supervision in target dataset, we can transfer discriminative knowledge from supervised to unsupervised dataset.

With the goal of reducing information loss in the above procedure for compressing the visual feature maps, here we consider a decoder $D_C$ to reconstruct visual feature maps $X^t$/$X^s$ from the compact domain-invariant and specific features $(e^t_c,e^t_p)$/$(e^s_c,e^s_p)$. Thus, we define reconstruction loss $\mathcal{L}_{rec}$ as:



\begin{equation}\label{eq:rec}
\begin{aligned}
\mathcal{L}_{rec} = \sum_{i=1}^{N_s} {\lVert X_i^{s} - \hat{X}_i^{s}\rVert}_2^{2} + \sum_{i=1}^{N_t} {\lVert X_i^{t} - \hat{X}_i^{t}\rVert}_2^{2}
\end{aligned}
\end{equation}
where $X^s_i$/$X^t_i$ and $\hat{X}^s_i$/$\hat{X}^t_i$ denote encoded and reconstructed the visual feature maps for source/target dataset respectively.




Note that the above learning objectives cannot ensure that the dataset-invariant and specific feature are mutual exclusive and independent, we therefore introduce a difference loss $\mathcal{L}_{diff}$ to encourage the orthogonality between these two features:
\begin{equation}\label{eq:mse}
\begin{aligned}
\mathcal{L}_{diff}= {\lVert {\textbf{H}_c^s}^\top \textbf{H}_p^s \rVert}_{F}^{2} +  {\lVert {\textbf{H}_c^t}^\top \textbf{H}_p^t \rVert}_{F}^{2}
\end{aligned}
\end{equation}
where $\textbf{H}_c^s$ and $\textbf{H}_c^t$ be matrices whose rows are the latent \textit{shared} representations $e_c^s = E_C(X^s)$ and $e_c^t = E_C(X^t)$. $\textbf{H}_p^s$ and $\textbf{H}_p^t$ are obtained in a similar manner. Note that ${\lVert \cdot \rVert}_{F}^{2}$ is the square Frobenius norm.

\subsection{Learning and Performing Re-ID}\label{ssec:learn_eva}
\begin{figure}[t!]
	\centering
	\includegraphics[width=1\textwidth]{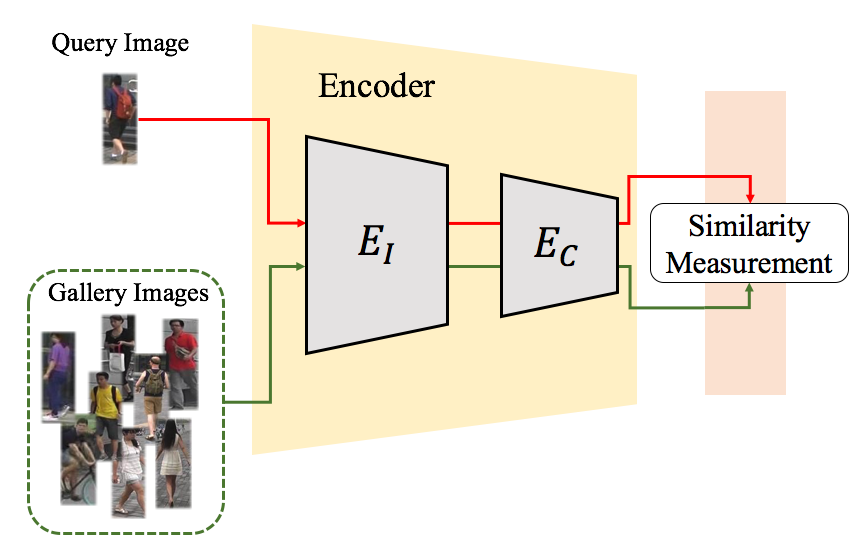}
	
     \caption{When performing Re-ID using our ARN, only $E_I$ and $E_c$ in the latent encoder are required. That is, we match the latent feature $e^q_c$ of person-of-interest (in the dataset) with the latent feature $e^t_c$ of the test image by calculating the similarity ranking.}
	\vspace{0mm}
	\label{fig:test}
\end{figure}
In sum, the total training objective $\mathcal{L}_{total}$ for our ARN can be written as follows:
\small
\begin{equation}\label{eq:euclidean}
\begin{aligned}
\mathcal{L}_{total} = \mathcal{L}_{class} + \alpha \cdot \mathcal{L}_{ctrs}+ \beta \cdot \mathcal{L}_{rec}+\gamma \cdot\mathcal{L}_{diff}
\end{aligned}
\end{equation}
\normalsize
where $\alpha$, $\beta$, and $\gamma$ are hyper-parameters that control the interaction of the total loss. We train our model by the minimizing $\mathcal{L}_{total}$ in an end-to-end manner. Once the model is learned, as shown in Figure~\ref{fig:test}, our model performs Re-ID by measuring the cosine similarity of features of query and gallery images. 

Note that our ARN is able to perform Re-ID task in the unsupervised dataset by adapting discriminative ability from source to target domain.

\section{Experiments} \label{sec:results}
{
We now evaluate the performance of our proposed network, which is applied to perform cross-domain Re-ID tasks. 
To verify the work of each component in ARN, we provide ablation studies in Section~\ref{sec:abl}. Furthermore, in Section~\ref{sec:sta}, we compare the performance of our ARN with several supervised and unsupervised methods.


}

\subsection{Datasets}\label{sec:dataset}

	To evaluate our proposed method, we conduct experiments on Market-1501~\cite{zheng2015scalable} and DukeMTMC-reID~\cite{zheng2017unlabeled}, because both datasets are large-scale and commonly used. The details of the number of training samples under each camera are shown in Table.~\ref{tab:stat}. 

\textbf{Market-1501}~\cite{zheng2015scalable} is composed of  32,668 labeled images of 1,501 identities collected from 6 camera views. The dataset is split into two non-over-lapping fixed parts: 12,936 images from 751 identities for training and 19,732 images from 750 identities for testing. In testing, 3368 query images from 750 identities are used to retrieve the matching persons in the gallery. 


\textbf{DukeMTMC-reID}~\cite{zheng2017unlabeled} is also a large-scale Re-ID dataset. It is collected from 8 cameras and contains 36,411 labeled images belonging to 1,404 identities. It also consists of 16,522 training images from 702 identities, 2,228 query images from the other 702 identities, and 17,661 gallery images. 

We use rank-1 accuracy and mean average precision (mAP) for evaluation on both datasets. In the experiments, there are two source-target settings: 
\begin{enumerate}
\item Target: Market-1501 / Source: DukeMTMC-reID.
\item Target: DukeMTMC-reID / Source: Market-1501.
\end{enumerate}

\begin{table}[t]
\centering
\caption{Numbers of training samples and cameras in Market-1501 and DukeMTMC-reID datasets.}
\label{tab:stat}
\begin{tabular}{l|l|l|l}
\hline
\multicolumn{2}{l|}{Market-1501} & \multicolumn{2}{l}{DukeMTMC-reID} \\ \hline
camera       & \# of images       & camera        & \# of images       \\ \hline
1            & 2017               & 1             & 2809               \\ 
2            & 1709               & 2             & 3009               \\ 
3            & 2707               & 3             & 1088               \\ 
4            & 920                & 4             & 1395               \\ 
5            & 2338               & 5             & 1685               \\ 
6            & 3245               & 6             & 3700               \\ 
             &                    & 7             & 1330               \\ 
             &                    & 8             & 1506               \\ \hline
\end{tabular}
\end{table}
\subsection{Implementation Details}\label{sec:imp}
\begin{table*}[t]
\centering
\caption{Ablation studies of Adaptation and Re-Identification Network (ARN) under different experimental settings.}
\label{tab:var}
\begin{tabular}{l|l l l l l| l l l l l}
\hline
\multirow{2}{*}{Method}              & \multicolumn{5}{l|}{\begin{tabular}[c]{@{}l@{}}Target: Market-1501 \\ Source: DukeMTMC-reID\end{tabular}} & \multicolumn{5}{l}{\begin{tabular}[c]{@{}l@{}}Target: DukeMTMC-reID\\ Source: Market-1501\end{tabular}} \\ \cline{2-11} 
                                     & R1                 & R5                 & R10                 & R20                 & mAP                 & R1                 & R5                 & R10                 & R20                 & mAP                \\ \hline\hline
Ours w/o $\mathcal{L}_{ctrs}$, $\mathcal{L}_{class}$, $E_S$, $E_T$  &44.5& 63.2 & 70.4 & 78.5 & 20.3& 31.2& 42.5 &50.1 &57.4&18.4\\ 

Ours w/o $\mathcal{L}_{ctrs}$, $\mathcal{L}_{class}$  &52.2 & 68.4 & 75.9 &82.1 & 23.7 & 36.7&48.9 & 58.2 &63.4 &19.6\\
Ours w/o $E_S$, $E_T$ &60.5 & 74.2 & 81.9 &88.1 & 28.7 & 48.4 & 62.5&68.8 &73.1& 26.8 \\ 
Ours &  \textbf{70.3}   & \textbf{80.4}  & \textbf{86.3} & \textbf{93.6}  &\textbf{39.4} &  \textbf{60.2}  & \textbf{73.9}  & \textbf{79.5} & \textbf{82.5} & \textbf{33.4}\\ \hline
\end{tabular}
\end{table*}

\textbf{ARN}.
Following Section~\ref{sec:method}, we use ResNet-50 pre-trained on ImageNet as our $E_I$ model in the encoder. In order to perform the latent embedding easily for the modules $E_T$, $E_C$, and $E_S$, we remove the last few layers including average pooling from the pre-trained ResNet-50 model. The input of the $E_I$ will be images with size $224 \times 224 \times 3$, denoting width, height, and channel respectively. In this manner, the output of $E_I$ is the feature-map $X$ with size $7 \times 7 \times 2048$ and will be fed into $E_T$, $E_C$, and $E_S$ to obtain the corresponding feature with size $1 \times 1 \times 2048$, which is then flatten to private $e_p$ or sharefji32l4d $e_c$ latent feature with size $2048$ as the final output of the encoder. Note that $E_T$, $E_C$, and $E_S$ are implemented with fully convolution networks (FCNs) which contains three layers. 

The input of our decoder $D_c$ is the concatenated latent feature $(e_c, e_p)$ with size $4096$. We also implement the latent decoder $D_c$ with fully convolution network. The output size of the decoder $D_c$ is $7 \times 7 \times 2048$, which is identical to the input of $E_T$, $E_C$, and $E_S$ modules. Note that the concatenated vectors in both domains, $(e_c^t,e_p^t)$ and $(e_c^s,e_p^s)$, are fed into the latent decoder simultaneously during the training procedure. 

The classifier $C_S$ contains only fully connected layers with dropout mechanism. We only feed the shared latent feature $e_c^s$ into the classifier. The output is the classification result among the identities. That is, the output size would be 702 if the source domain is DukeMTMC-reID~\cite{zheng2017unlabeled} or 751 if the source domain is Market-1501~\cite{zheng2015scalable}.

\textbf{Learning procedure}.
As mentioned in Section~\ref{sec:method}, we aim to minimize the total loss $\mathcal{L}_{total}$ in Equation~\ref{eq:euclidean} during the training procedure. The parameters $\alpha$, $\beta$, and $\gamma$ are chosen under the experimental trials. In practice, we set $\alpha$, $\beta$, and $\gamma$ as $0.01$, $2.0$, and $1500$, respectively. We aim to balance the larger value of $\mathcal{L}_{ctrs}$ and the smaller one of $\mathcal{L}_{diff}$. In addition, we need larger weight to enforce the reconstruction.



While we can directly use the same learning rate for each component to update the whole network, it might result in overfitting issues.
We believe that individually setting the customized learning rates for $E_I$, $E_T$, $E_C$, $E_S$, $D_C$, and $C_S$ can avoid this problem. For instance, when minimizing $\mathcal{L}_{class}$ and $\mathcal{L}_{ctrs}$, the weights of the pre-trained model $E_I$ should not be updated faster than other modules because we try to keep much useful pre-trained weights ever trained on ImageNet. Hence, we set the learning rate for $E_I$ to a relatively small value, $10^{-7}$, and only tune $E_I$ in the first few epochs. In addition, we set the learning rate of $E_T$, $E_C$, $E_S$, $D_C$ to $10^{-3}$, and $C_S$ to $2 \times 10^{-3}$. We adopt the stochastic gradient descent (SGD) to update the parameters of the network. 


	

\textbf{Evaluating procedure}.
At the end of the learning scenario, we proceed to evaluate the performance of our trained network on Re-ID task.
We only use $E_I$ and $E_c$ in the encoder for generating the latent features in evaluating scenario. For performance evaluation, we sort the cosine distance between the query and all the gallery features to obtain the final ranking result. Note that the cosine distance is equivalent to Euclidean distance when the feature is L2-normalized. Moreover, we employ the standard metrics as in most person Re-ID literature, namely the cumulative matching curve (CMC) used for generating ranking accuracy, and the mean Average Precision (mAP).

\subsection{Ablation Studies}\label{sec:abl}

In this subsection, we aim to fully analyze the effectiveness of our ARN via comparing with other baseline settings. As shown in Table~\ref{tab:var}, we compare our final version model with the ones removing supervised losses $\mathcal{L}_{ctrs}$, $\mathcal{L}_{class}$ in source domain or private components $E_S$, $E_T$. For dataset Market-1501 and DukeMTMC-reID, our full model can achieve $70.3\%$ and $60.2\%$ at Rank-1 accuracy, and $39.4\%$ and $33.4\%$ at mAP respectively.

\textbf{Reconstruction loss $\mathcal{L}_{rec}$}.
For the target dataset on Market-1501 and DukeMTMC-reID, we observe that the Rank-1 accuracy of baseline model without $\mathcal{L}_{ctrs}$, $\mathcal{L}_{class}$, $E_S$, and $E_T$, containing only the reconstruction loss $\mathcal{L}_{rec}$, decrease by $25.8\%$ and by $29\%$ respectively. However, this shows that the reconstruction loss does play a great role in learning basic latent representation, which can still achieve $44.5\%$ and $31.2\%$ at Rank-1 accuracy. We note that without $\mathcal{L}_{ctrs}$, $\mathcal{L}_{class}$, we are not able to fine-tune $E_I$ and let $E_I$ keep its original pre-trained weights on ImageNet.

\textbf{Source supervised losses $\mathcal{L}_{ctrs}, \mathcal{L}_{class}$}. Refer to Table~\ref{tab:var} again, we also observe that without supervised loss $\mathcal{L}_{ctrs}, \mathcal{L}_{class}$, the Rank-1 accuracy decrease by $18.1\%$ and by $21.4\%$ on Market-1501 and DukeMTMC-reID respectively. This obvious drop indicates that supervised metrics on source domain has largely improved the performance of our ARN model. We also conclude that the shared latent space does need the losses $\mathcal{L}_{ctrs}, \mathcal{L}_{class}$ to capture the semantics of person information.

\textbf{Private modules $E_T$, $E_S$}. In Table~\ref{tab:var}, without private modules $E_T$, $E_S$, the Rank-1 accuracy decrease by $9.8\%$ and by $11.8\%$ on Market-1501 and DukeMTMC-reID respectively. We conclude that without partitioning the space to produce a private representation, the feature space may be contaminated with aspects of the noise that are unique for each dataset. Hence, having the private modules $E_T$, $E_S$ does help perform representation learning in the shared latent space.


\begin{table}[t]
\centering
\caption{Performance comparisons on Market-1501 with supervised and unsupervised Re-ID methods.}
\label{tab:market}
\begin{tabular}{ l|l|c c c c }
\hline
&Method              & Rank-1 & Rank-5 & Rank-10 & mAP         \\ \hline\hline
\parbox[t]{2mm}{\multirow{6}{*}{\rotatebox[origin=c]{90}{Supervised}}}
&BOW~\cite{zheng2015scalable}                 & 44.4  &-&-& 20.8  \\ 
&LDNS~\cite{zhang2016learning}                & 61.0  &-&-&35.7   \\ 
&SVDNET~\cite{sun2017svdnet}              & 82.3   &-&-& 62.1    \\ 
&TriNet~\cite{hermans2017defense}              & 84.9  &-&-& 69.1  \\ 
&CamStyle~\cite{zhong2017camera}            & 89.5  &-&-& 71.6   \\ 
&DuATM~\cite{si2018dual}               & 91.4  &-&-& 76.6 \\ \hline\hline
\parbox[t]{2mm}{\multirow{6}{*}{\rotatebox[origin=c]{90}{Unsupervised}}}
&BOW~\cite{zheng2015scalable}                 & 35.8   & 52.4   & 60.3    & 14.8  \\ 
&UMDL~\cite{peng2016unsupervised}                & 34.5   & 52.6   & 59.6    & 12.4  \\ 
&PUL~\cite{fan2017unsupervised}                 & 45.5   & 60.7   & 66.7    & 20.5  \\ 
&CAMEL~\cite{yu2017cross}               & 54.5   & -      & -       & 26.3    \\ 
&SPGAN~\cite{image-image18}               & 57.7   & 75.8   & 82.4    & 26.7   \\ 
&\textbf{Ours}				& \textbf{70.3}   & \textbf{80.4}  & \textbf{86.3}    &\textbf{39.4}           \\ \hline
\end{tabular}
\end{table}

\subsection{Comparison with State-of-the-art Methods}\label{sec:sta}

~\textbf{Market-1501.} In Table~\ref{tab:market}, we first compare our model with the unsupervised methods. For the hand-crafted features based models, we compare our model with Bag-of-Word (BOW)~\cite{zheng2015scalable}. 
For the cross-domain Re-ID models, there are Unsupervised Multi-task Dictionary Learning (UMDL)~\cite{peng2016unsupervised}, Progressive Unsupervised Learning (PUL)~\cite{fan2017unsupervised}, Clustering-based Asymmetric Metric Learning (CAMEL)~\cite{yu2017cross} and Similarity Preserving Generative Adversarial Network (SPGAN)~\cite{image-image18}. Our model outperforms these models in Rank-1, Rank-5, Rank-10, and mAP on Market-1501. Note that our model outperforms the second best method by $13.6\%$ in Rank-1 accuracy and by $12.7\%$ in mAP. 

\begin{table}[t]
\centering
\caption{Performance comparisons on DukeMTMC-reID with supervised and unsupervised Re-ID methods.}
\label{tab:duke}
\begin{tabular}{l|l|c c c c }
\hline
&Method              & Rank-1 & Rank-5 & Rank-10 & mAP   \\ \hline\hline
\parbox[t]{2mm}{\multirow{6}{*}{\rotatebox[origin=c]{90}{Supervised}}}
&BOW~\cite{zheng2015scalable}                 & 25.1  &-&-& 12.2 \\
&LOMO~\cite{liao2015person}                & 30.8  &-&-& 17.0  \\ 
&TriNet~\cite{hermans2017defense}              & 72.4  &-&-& 53.5 \\ 
&SVDNET~\cite{sun2017svdnet}              & 76.7   &-&-& 56.8  \\ 
&CamStyle~\cite{zhong2017camera}            & 78.3  &-&-& 57.6 \\ 
&DuATM~\cite{si2018dual}               & 81.8  &-&-& 64.6\\ \hline \hline
\parbox[t]{2mm}{\multirow{5}{*}{\rotatebox[origin=c]{90}{Unsupervised}}}
&BOW~\cite{zheng2015scalable}                 & 17.1  & 28.8   & 34.9    & 8.3\\ 
&UMDL~\cite{peng2016unsupervised}                & 18.5   & 31.4   & 37.6    & 7.3  \\ 
&PUL~\cite{fan2017unsupervised}                 & 30.0     & 43.4   & 48.5    & 16.4 \\ 
&SPGAN~\cite{image-image18}               & 46.4   & 62.3   & 68.0      & 26.2 \\ 
&\textbf{Ours} & \textbf{60.2}  & \textbf{73.9}  & \textbf{79.5}   & \textbf{33.4}            \\ \hline
\end{tabular}
\end{table}

In addition, we also compare our model with existing supervised models, observing that our model surpasses BOW~\cite{zheng2015scalable}, LDNS~\cite{zhang2016learning} and already boost the performance closely to supervised deep learning based model like SVDNET~\cite{sun2017svdnet}, TriNet~\cite{hermans2017defense}, CamStyle~\cite{zhong2017camera}, or DuATM~\cite{si2018dual}.

~\textbf{DukeMTMC-reID.} In Table~\ref{tab:duke}, our model outperforms unsupervised methods such as BOW~\cite{zheng2015scalable}, UMDL~\cite{peng2016unsupervised}, PUL~\cite{fan2017unsupervised}, and SPGAN~\cite{image-image18}. Our model achieves~\textbf{Rank-1 accuracy=60.2\% and mAP=33.4\%} and outperforms the second best method~\cite{image-image18} roughly by $13.8\%$ in Rank-1 accuracy and by $7.2\%$ in mAP. More importantly, the performance of our model is better than some supervised methods such as BOW~\cite{zheng2015scalable} and LOMO~\cite{liao2015person}.

\vspace{3mm}

\section{Conclusions} \label{sec:conclusions}
In this paper, we presented a deep learning model of Adaptation and Re-Identification Network (ARN) for solving cross-domain Re-ID tasks. Our ARN allows us to jointly exploit a pre-collected supervised source-domain dataset and a target-domain dataset of interest by learning domain invariant and discriminative features. As a result, Re-ID in the target-domain can be performed even without any label information observed during training. With this proposed unsupervised domain adaptation network, we conducted experiments on Market-1501 and DukeMTMC-reID datasets, and confirmed the effectiveness of our model in such a challenging unsupervised learning setting. Moreover, our method also performed favorably against a number of baseline supervised Re-ID approaches, which again supports the use of our ARN for practical Re-ID tasks.

\noindent\textbf{Acknowledgments}
This work was supported in part by Umbo CV and the Ministry of Science and Technology of Taiwan under grant MOST 107-2634-F-002-010.

{\small
\bibliographystyle{ieee}
\bibliography{egbib}

\begin{thebibliography}{10}\itemsep=-1pt

\bibitem{bousmalis2016domain}
K.~Bousmalis, G.~Trigeorgis, N.~Silberman, D.~Krishnan, and D.~Erhan.
\newblock Domain separation networks.
\newblock In {\em Advances in Neural Information Processing Systems (NIPS)},
  pages 343--351, 2016.

\bibitem{cheng2016person}
D.~Cheng, Y.~Gong, S.~Zhou, J.~Wang, and N.~Zheng.
\newblock Person re-identification by multi-channel parts-based cnn with
  improved triplet loss function.
\newblock In {\em Proceedings of the IEEE Conference on Computer Vision and
  Pattern Recognition (CVPR)}, 2016.

\bibitem{image-image18}
W.~Deng, L.~Zheng, Q.~Ye, G.~Kang, Y.~Yang, and J.~Jiao.
\newblock Image-image domain adaptation with preserved self-similarity and
  domain-dissimilarity for person re-identification.
\newblock In {\em Proceedings of the IEEE Conference on Computer Vision and
  Pattern Recognition (CVPR)}, 2018.

\bibitem{fan2017unsupervised}
H.~Fan, L.~Zheng, and Y.~Yang.
\newblock Unsupervised person re-identification: Clustering and fine-tuning.
\newblock In {\em arXiv preprint}, 2017.

\bibitem{geng2016deep}
M.~Geng, Y.~Wang, T.~Xiang, and Y.~Tian.
\newblock Deep transfer learning for person re-identification.
\newblock In {\em arXiv preprint arXiv:1611.05244}, 2016.

\bibitem{hadsell2006dimensionality}
R.~Hadsell, S.~Chopra, and Y.~LeCun.
\newblock Dimensionality reduction by learning an invariant mapping.
\newblock In {\em Proceedings of the IEEE Conference on Computer Vision and
  Pattern Recognition (CVPR)}, volume~2, pages 1735--1742. IEEE, 2006.

\bibitem{hermans2017defense}
A.~Hermans, L.~Beyer, and B.~Leibe.
\newblock In defense of the triplet loss for person re-identification.
\newblock In {\em arXiv preprint}, 2017.

\bibitem{liao2015person}
S.~Liao, Y.~Hu, X.~Zhu, and S.~Z. Li.
\newblock Person re-identification by local maximal occurrence representation
  and metric learning.
\newblock In {\em Proceedings of the IEEE Conference on Computer Vision and
  Pattern Recognition (CVPR)}, pages 2197--2206, 2015.

\bibitem{lin2017improving}
Y.~Lin, L.~Zheng, Z.~Zheng, Y.~Wu, and Y.~Yang.
\newblock Improving person re-identification by attribute and identity
  learning.
\newblock In {\em arXiv preprint}, 2017.

\bibitem{liu2017unsupervised}
M.-Y. Liu, T.~Breuel, and J.~Kautz.
\newblock Unsupervised image-to-image translation networks.
\newblock In {\em Advances in Neural Information Processing Systems (NIPS)},
  2017.

\bibitem{liu2016coupled}
M.-Y. Liu and O.~Tuzel.
\newblock Coupled generative adversarial networks.
\newblock In {\em Advances in Neural Information Processing Systems (NIPS)},
  2016.

\bibitem{patel2015visual}
V.~M. Patel, R.~Gopalan, R.~Li, and R.~Chellappa.
\newblock Visual domain adaptation: A survey of recent advances.
\newblock {\em IEEE Signal Processing Magazine}, 32(3):53--69, 2015.

\bibitem{peng2016unsupervised}
P.~Peng, T.~Xiang, Y.~Wang, M.~Pontil, S.~Gong, T.~Huang, and Y.~Tian.
\newblock Unsupervised cross-dataset transfer learning for person
  re-identification.
\newblock In {\em Proceedings of the IEEE Conference on Computer Vision and
  Pattern Recognition (CVPR)}, pages 1306--1315, 2016.

\bibitem{si2018dual}
J.~Si, H.~Zhang, C.-G. Li, J.~Kuen, X.~Kong, A.~C. Kot, and G.~Wang.
\newblock Dual attention matching network for context-aware feature sequence
  based person re-identification.
\newblock In {\em arXiv preprint}, 2018.

\bibitem{sun2017svdnet}
Y.~Sun, L.~Zheng, W.~Deng, and S.~Wang.
\newblock Svdnet for pedestrian retrieval.
\newblock In {\em arXiv preprint}, 2017.

\bibitem{tzeng2014deep}
E.~Tzeng, J.~Hoffman, N.~Zhang, K.~Saenko, and T.~Darrell.
\newblock Deep domain confusion: Maximizing for domain invariance.
\newblock {\em arXiv preprint}, 2014.

\bibitem{wang2016towards}
H.~Wang, X.~Zhu, T.~Xiang, and S.~Gong.
\newblock Towards unsupervised open-set person re-identification.
\newblock In {\em Proceedings of the IEEE International Conference on Image
  Processing (ICIP)}. IEEE, 2016.

\bibitem{yu2017cross}
H.-X. Yu, A.~Wu, and W.-S. Zheng.
\newblock Cross-view asymmetric metric learning for unsupervised person
  re-identification.
\newblock In {\em Proceedings of the IEEE International Conference on Computer
  Vision (ICCV)}, 2017.

\bibitem{zhang2016learning}
L.~Zhang, T.~Xiang, and S.~Gong.
\newblock Learning a discriminative null space for person re-identification.
\newblock In {\em Proceedings of the IEEE Conference on Computer Vision and
  Pattern Recognition (CVPR)}, 2016.

\bibitem{zheng2015scalable}
L.~Zheng, L.~Shen, L.~Tian, S.~Wang, J.~Wang, and Q.~Tian.
\newblock Scalable person re-identification: A benchmark.
\newblock In {\em Proceedings of the IEEE International Conference on Computer
  Vision (ICCV)}, 2015.

\bibitem{zheng2016person}
L.~Zheng, Y.~Yang, and A.~G. Hauptmann.
\newblock Person re-identification: Past, present and future.
\newblock In {\em arXiv preprint}, 2016.

\bibitem{zheng2017unlabeled}
Z.~Zheng, L.~Zheng, and Y.~Yang.
\newblock Unlabeled samples generated by gan improve the person
  re-identification baseline in vitro.
\newblock In {\em Proceedings of the IEEE International Conference on Computer
  Vision (ICCV)}, 2017.

\bibitem{zhong2017camera}
Z.~Zhong, L.~Zheng, Z.~Zheng, S.~Li, and Y.~Yang.
\newblock Camera style adaptation for person re-identification.
\newblock In {\em arXiv preprint}, 2017.

\bibitem{CycleGAN2017}
J.-Y. Zhu, T.~Park, P.~Isola, and A.~A. Efros.
\newblock Unpaired image-to-image translation using cycle-consistent
  adversarial networkss.
\newblock In {\em Proceedings of the IEEE International Conference on Computer
  Vision (ICCV)}, 2017.

\end{thebibliography}
}

\end{document}